\documentclass{article}

\usepackage{PRIMEarxiv}

\usepackage[utf8]{inputenc} % allow utf-8 input
\usepackage[T1]{fontenc}    % use 8-bit T1 fonts
\usepackage{hyperref}       % hyperlinks
\usepackage{url}            % simple URL typesetting
\usepackage{booktabs}       % professional-quality tables
\usepackage{amsfonts}       % blackboard math symbols
\usepackage{nicefrac}       % compact symbols for 1/2, etc.
\usepackage{microtype}      % microtypography
\usepackage{lipsum}
\usepackage{fancyhdr}       % header
\usepackage{graphicx}       % graphics
\graphicspath{{media/}}     % organize your images and other figures under media/ folder

\usepackage{latexsym}
\usepackage{graphicx}
\usepackage{multicol,multirow}
\usepackage{amsmath,amssymb,amsfonts}
\usepackage{mathrsfs}
\usepackage{amsthm}
\usepackage{rotating}
\usepackage{appendix}
\usepackage{ifpdf}
\usepackage[T1]{fontenc}
\usepackage{times}
\usepackage{newtxmath}
\usepackage{textcomp}%
\usepackage{xcolor}%
\usepackage{hyperref}
\usepackage{lipsum}
\usepackage{authblk}

\title{Physics-Informed Learning of Aerosol Microphysics}

\author[1, 2, 3, 4]{Paula Harder}
\author[1]{Duncan Watson-Parris}
\author[1]{Philip Stier}
\author[2]{Dominik Strassel}
\author[4]{Nicolas R. Gauger}
\author[2,3,5]{Janis Keuper}

\affil[1]{Atmospheric, Oceanic and Planetary Physics, Department of Physics, Universty of Oxford}
\affil[2]{Fraunhofer Center High-Performance Computing, Fraunhofer ITWM}
\affil[3]{Fraunhofer Center Machine Learning, Fraunhofer Society}
\affil[4]{Chair for Scientific Computing, TU Kaiserslautern}
\affil[5]{Institute for Machine Learning and Analytics, Institute for Machine Learning and Analytics}

\begin{document}
\maketitle

Aerosol particles play an important role in the climate system by absorbing and scattering radiation and influencing cloud properties. They are also one of the biggest sources of uncertainty for climate modeling. Many climate models do not include aerosols in sufficient detail due to computational constraints. In order to represent key processes, aerosol microphysical properties and processes have to be accounted for. This is done in the ECHAM-HAM global climate aerosol model using the M7 microphysics, but high computational costs make it very expensive to run with finer resolution or for a longer time. We aim to use machine learning to emulate the microphysics model at sufficient accuracy and reduce the computational cost by being fast at inference time. The original M7 model is used to generate data of input-output pairs to train a neural network on it. We are able to learn the variables' tendencies achieving an average $R^2$ score of $77.1\% $. We further explore methods to inform and constrain the neural network with physical knowledge to reduce mass violation and enforce mass positivity. On a GPU we achieve a speed-up of up to over 64x compared to the original model.

\section{Introduction}
Aerosol forcing remains the largest source of uncertainty in the anthropogenic effect on the current climate \cite{https://doi.org/10.1029/2019RG000660}. The aerosol cooling effect hides some of the positive radiative forcing caused by greenhouse gas emissions and future restrictions to lower air pollution might result in stronger observed warming. Aerosols impact climate change through aerosol-radiation interactions and aerosol-cloud interactions \cite{IPCCWG1PhysicalStocker2013}. They can either scatter or absorb radiation, which depends on the particles' compounds. Black carbon aerosols from fossil fuel burning for example have a warming effect by strongly absorbing radiation, whereas sulphate from volcanic eruptions has a cooling effect by being less absorbing and primarily scattering radiation. Clouds influence the earth's radiation budget by reflecting sunlight and aerosols can change cloud properties significantly by acting as cloud condensation nuclei (CCN). A higher concentration of aerosols leads to more CCN which, for a fixed amount of water, results in more but smaller cloud droplets. Smaller droplets increase the clouds' albedo \cite{TWOMEY19741251} and can enhance the clouds' lifetime \cite{ALBRECHT1227}. 

Many climate models consider aerosols only as external parameters, they are read once by the model, but then kept constant throughout the whole model run. In the case where some aerosol properties are modeled, there might be no distinction between aerosol types, and just an overall mass is considered. To incorporate a more accurate description of aerosols, aerosol-climate modeling systems like ECHAM-HAM \cite{acp-5-1125-2005} have been introduced. It couples the ECHAM General Circulation Model (GCM) with a complex aerosol model called HAM. The microphysical core of HAM is either SALSA \cite{acp-8-2469-2008} or the M7 model \cite{m7}. We consider the latter here. M7 uses seven log-normal modes to describe aerosol properties, the particle sizes are represented by four size modes, nucleation, aitken, accumulation, and coarse, of which the aitken, accumulation, and coarse can be either soluble or insoluble \footnotetext[1]{In the paper we will use following abbreviations: NS=nucleation soluble, KS/KI=aitken soluble/insoluble, AS/AI=accumulation soluble/insoluble, CS/CI=coarse soluble/insoluble}. It includes processes like nucleation, coagulation, condensation, and water uptake, which lead to the redistribution of particle numbers and mass among the different modes. In addition, M7 considers five different components - Sea salt (SS), sulfate (SO4), black carbon (BC), primary organic carbon (OC), and dust (DU). M7 is applied to each grid box independently, it does not model any spatial relations.

\begin{figure}[t]
\begin{center}
\title{Change in H2SO4 concentration}
\vskip 0.1in
\centerline{\includegraphics[width=\columnwidth]{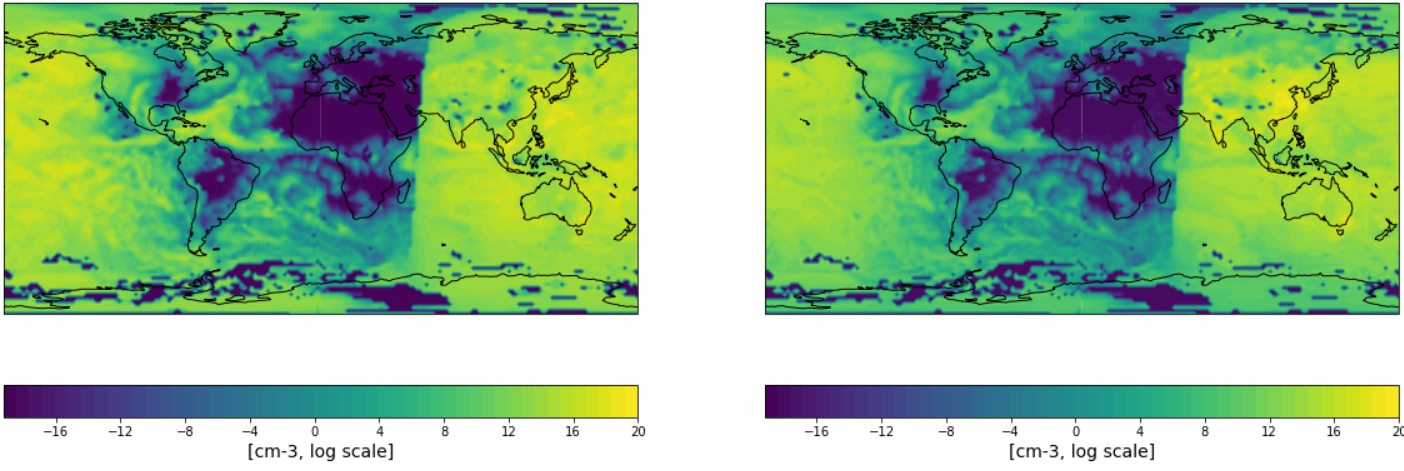}}
\caption{The change in concentration modeled by the M7 module for the first time step of the test data is plotted on the left. The predicted change is plotted on the right. Both plots show the change in concentration on a logarithmic scale}
\label{map_plot}
\end{center}
\vskip -0.2in
\end{figure} 

More detailed models come with the cost of increased computational time: ECHAM-HAM can be run at 150km resolution for multiple decades. But to run storm-resolving models, for example, the goal is ideally a 1km horizontal grid resolution and still be able to produce forecasts up to a few decades. If we want to keep detailed aerosol descriptions for this a significant speedup of the aerosol model is needed. 

Replacing climate model components with machine learning approaches and therefore decreasing a model's computing time has shown promising results in the past. There are several works on emulating convection, both random forest approaches, \cite{gorman} and deep learning models \cite{Rasp9684}, \cite{gentine}, \cite{beucler2020physicallyconsistent} have been explored. Recently, multiple consecutive neural networks have been used to emulate a bin microphysical model for warm rain processes \cite{gettelmann}. Silva et al. \cite{gmd-2020-393} compare several methods, including deep neural networks, XGBoost, and ridge regression as physically regularized emulators for aerosol activation. In addition to the aforementioned approaches random forest approaches have been used to successfully derive the CCN from atmospheric measurements \cite{acp-20-12853-2020}. Next to many application cases there now exist two benchmark datasets for climate model emulation \cite{10.1002/essoar.10509765.1}, \cite{cachay2021climart}.

Physics-Informed machine learning recently gained attention in machine learning research, including machine learning for weather and climate modeling \cite{climpinn}. Whereas a large amount of work focuses on soft-constraining neural networks by adding equations to the loss term, there was also the framework DC3 \cite{donti2021dc3} developed, that incorporates hard constraints. Using emulators in climate model runs can require certain physical constraints to be met, that are usually not automatically learned by a neural network. For convection emulation there has been work on both hard and soft constraints to enforce conversation laws, adding a loss term or an additional network layer \cite{beucler21}. 

Building on our previous work \cite{harder2021emulating}, we demonstrate a machine learning approach to emulate the M7 microphysics module. We investigated different approaches, neural networks as well as ensemble models like random forest and gradient boosting, with the aim of achieving the desired accuracy and computational efficiency, finding the neural network (NN) appeared to be the most successful. We use data generated from a realistic ECHAM-HAM simulation and train a model offline to predict one time step of the aerosol microphysics. The underlying data distribution is challenging, as the changes in the variables are often zero or very close to zero. We do not predict the full values, but the tendencies. An example of our global prediction is shown in Figure \ref{map_plot}. To incorporate physics into our network we explore both soft constraints for our emulator by adding regularization terms to the loss function and hard constraints by adding completion and correction layers.
\section{Methodology}
\subsection{Data}
\subsubsection{Data Generation}
To easily generate data we extract the aerosol microphysics model from the global climate model ECHAM-HAM and develop a stand-alone version. To obtain input data ECHAM-HAM is run for four days within a year. We use a horizontal resolution of 150km at the equator, overall 31 vertical levels, 96 latitudes, and 192 longitudes. A time step length of 450s is chosen. This yields a data set of over 100M points for each day. We only use a subset of 5 times a day, resulting in about 2.85M points per day. To test if our emulator generalizes well in an unknown setting, we use the days in January and April for training, a day in July for validation, and the October data for testing. Because the two sea salt variables only change in $0.2\%$ of the time as they undergo little microphysical changes we do not model them here. The masses and particle concentrations for different aerosol types and different modes are both inputs and outputs, the output being the value one time step later. Atmospheric state variables like pressure, temperature, and relative humidity are used as inputs only. The dataset is available at \cite{paula_harder_2022_5837936}. A full list of input and output values can be found in the supplementary material.
\subsubsection{Data distribution and transformation}
Compared to the full values the tendencies (changes) are very small, therefore we aim to predict the tendencies, not the full values, where it is possible. Depending on the variable we have very different size scales, but also a tendency for a specific variable may span several orders of magnitude. In some modes, variables can only grow, in some only decrease, and in others do both. Often the majority of tendencies are either zero or very close to zero, but a few values might be very high. The absolute tendencies are roughly log-normal distributed. A logarithmic transformation has been used in \cite{gettelmann} to give more weight to the prominent case, the smaller changes. This then transforms our inputs close to a normal distribution, which is favorable for the neural network to be trained on. On the other hand, a linear scale might be more representative, as it gives more weight to values with the largest actual physical impact. Results and a discussion for training with logarithmically transformed variables can be found in the appendix.

\subsection{General Network Architecture}
For emulating the M7 microphysics model we explored different machine learning approaches, including random forest regression, gradient boosting, and a simple neural network (see suppl.). Providing more expressivity, the neural network approach appears to be the most successful for this application and will be shown here. We do not only present one neural model here, but different versions building on the same base architecture and discuss the advantages and disadvantages of these different designs.

We employ a fully connected neural network, where the values are propagated from the input layer through multiple hidden to the output layer, using a combination of linear operations and non-linear activations. We use a ReLU activation function (a comparison of scores for different activation functions can be found in the supplementary material) and three hidden layers, each hidden layer containing 128 nodes. Using zero hidden layers results in linear regression and does not have the required expressivity for our task, one hidden layer already achieves a good performance, after two layers the model we could not see any significant further improvements. 

\subsubsection{Training}
We train all our NNs using the Adam optimizer with a learning rate of $10^{-3}$, a weight decay of $10^{-9}$, and a batch size of 256. Our objective to optimize during the training of the neural networks is specified in Equation \ref{loss}, using an MSE loss in every version and activating the additional loss terms depending on the version. WE train our model for 100 epochs. The training on a single NVIDIA Titan V GPU takes about 2 hours.

For the classification network, we use a similar setup as described above. As a loss function be use the Binary-Cross-Entropy loss, the training takes about 30 minutes.

\subsection{Physics-Constrained Learning}

In this work, we explore two different ways of physics-informing similar to \cite{beucler21}: Adding regularizer terms to the loss function and adding hard constraints as an additional network layer. The physical constraints that naturally come up for our setting are mass conservation, the mass of one aerosol species has to stay constant, and mass or concentration positivity, a mass or concentration resulting from our predicted tendencies has to be positive. 

\subsubsection{Soft Constraining}

To encourage the neural network to decrease mass violation or negative mass/concentration values we add additional loss terms to the objective:

\begin{equation}
   \min_\theta \mathcal{L} (\Tilde{y},y_\theta) + \lambda\mathcal{L}^{\text{(mass)}} (y_\theta) + \mu\mathcal{L}^{\text{(pos)}} (y_\theta).
\end{equation}

Where $y_\theta = f_\theta (x)$ is the network output for an input $x\in \mathbb{R}^{32}$, parameterized by the network's weights $\theta$. In our case $\mathcal{L}$ is a MSE error, $\mu,\ \lambda\in \{ 0,1 \}$ depending if a regularizer is activated or not. The term $\mathcal{L}^{\text{(mass)}}$ penalizes mass violation:

\begin{equation}\label{loss}
    \mathcal{L}^{\text{(mass)}} (y_\theta) := \sum_{s \in S} \alpha_s|\sum_{i \in I_s} y_\theta^{(i)} |,
\end{equation}

 Here $S$ is the set of different species and $I_s$ are the indices for a specific species $s$. The parameters $\alpha_i$ need to be tuned, chosen too small the mass violation will not be decreased, if a factor is too large, the network can find the local optimum that is constantly zero, the values can be found in the supplementary material.
The other term penalizes negative mass values and is given by \footnotetext[2]{Skipping the back transformation in to original scale for a clearly notation, details on that are provided in the supplementary material.}:

\begin{equation}
    \mathcal{L}^{\text{(pos)}} (y_\theta) := \sum_{i=0}^{n-1}\beta_i \text{ReLU}(- (y_\theta^{(i)}+x_i))^2
\end{equation}

Using a ReLU function, all negative values are getting penalized. In our case $n=28$, $x_i$ is the corresponding full input variable to predicted tendency $y_\theta^{(i)}$. The factors $\beta_i$ need to be tuned, similar to $\alpha_i$.

\subsubsection{Hard Constraining}

In order to have a guarantee for mass conservation or positive masses, we implement a correction and a completion method. This is done at test time as an additional network layer, a version where this is implemented within the training of the neural network did not improve the performance. 
\paragraph{Inequality Correction}
For the correction method, predicted values that result in an overall negative mass or number are set to a value such that the full value is zero. For the neural networks intermediate output $ \tilde{y}_\theta$, the final output is given by:

\begin{equation}
     y_\theta^{(i)} = \text{ReLU}((\tilde{y}_\theta^{(i)}+x_i)) - x_i
\end{equation}

\paragraph{Equality Completion}
The completion method addresses mass violation, for each species one variable's prediction is replaced by the negative sum of the other same species' tendencies. We obtained the best results by replacing the worst-performing variable. This completion results in exact mass conservation. For a species $S$ we choose index $j\in I_S$, the completion layer is defined as follows:

\begin{equation}
     y_\theta^{(j)} = -\sum_{i \in I_s\setminus \{ j \}} y_\theta^{(i)}
\end{equation}

\section{Results}

\subsection{Predictive Performance}
\subsubsection{Metrics}

We consider multiple metrics to get an insight into the performance of the different emulator designs, covering overall predictive accuracy, mass violation, and predicting nonphysical values. We look at the $R^2$ score and the MSE. To understand mass violation in we look at the mass biases for the different species and the overall mass violation, where all scores are normalized by the mean over the respective species. The metrics are completed with two scores about negative value predictions: An overall fraction of negative and therefore nonphysical predictions and the average negative extend per predicted value. For all the different scores and architectures we take the mean over five different random initializations of the underlying neural network.

\subsubsection{Evaluation and Comparison}

\begin{table}[htb]
\tabcolsep=0pt%
{\caption{Test metrics for different architectures and transformations. Base means the usage of only the base NN, correct adds the correction procedure and complete the completion procedure. Mass reg. and positivity reg. include the regularization terms. Best scores are bold and second best scores are highlighted in blue .\label{tab2}}}
{
\begin{tabular*}{\textwidth}{@{\extracolsep{\fill}}lccccc@{}}\toprule%

{Architecture} & {Base} & {+Correct} & {+Complete} & {+Mass Loss} & {+Positivity Loss} \\\midrule
{R2}&\textcolor{blue}{$0.763$}&\textbf{0.771} &$0.738$&0.730 &0.709\\

{MSE}&\textcolor{blue}{0.162 }&\textbf{0.161} &\textcolor{blue}{0.162}&0.187 & 0.211\\

{Mass Bias SO4}&$1.1\cdot 10^{-5}$ &$8.5\cdot 10^{-5}$ &\textbf{0.00} & \textcolor{blue}{$8.6\cdot 10^{-6}$}& $1.0\cdot 10^{-3}$\\
{Mass Bias BC}&{$3.8\cdot 10^{-5}$} &$1.4\cdot 10^{-4}$ &\textbf{0.00}& \textcolor{blue}{$3.4\cdot 10^{-5}$}&$3.6\cdot 10^{-4}$\\
{Mass Bias OC}&{$3.3\cdot 10^{-5}$} &$6.0\cdot 10^{-5}$&\textbf{0.00}& \textcolor{blue}{$1.1\cdot 10^{-5}$}& $6.4\cdot 10^{-4}$\\
{Mass Bias DU}&{$1.0\cdot 10^{-6}$}&$3.9\cdot 10^{-5}$ &\textbf{0.00}& \textcolor{blue}{$2.8\cdot 10^{-7}$}& $1.5\cdot 10^{-5}$\\
{Mass Violation}&$3.7\cdot 10^{-4}$ &$1.1\cdot 10^{-3}$ &\textbf{0.00}& \textcolor{blue}{$1.4\cdot 10^{-4}$} & $2.4\cdot 10^{-3}$\\
{Neg. Fraction}&0.134 &\textbf{0.00} &0.146 & 0.144& \textcolor{blue}{0.0894}\\
{{Neg. Mean}}&0.00151&\textbf{0.00} &0.00170 & {0.00169}& \textcolor{blue}{0.000081} \\

\label{many_scores}
\end{tabular*}%
}
%\vskip -0.4in
\end{table}

\begin{figure}[htb]
\begin{center}
%\title{Change in H2SO4 concentration}
\vskip 0.1in
\centerline{\includegraphics[width=\columnwidth]{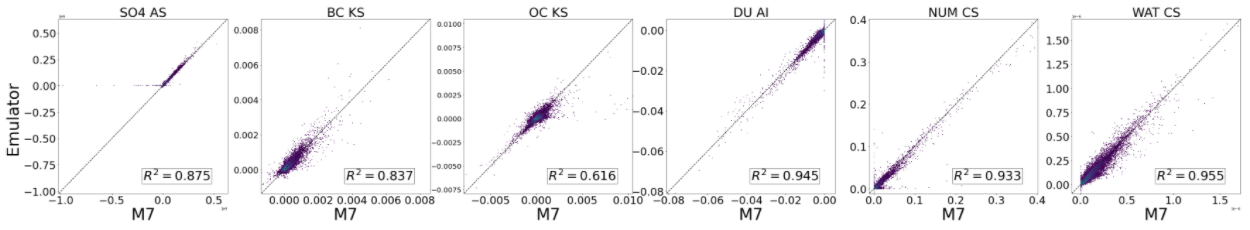}}
\caption{This figure shows the test prediction of our emulators against the true M7 values.For each type (species, number particles, water) we plot the performance of one variable (using the median or worse performing, see all in suppl.)}
\label{four}
\end{center}
\vskip -0.2in
\end{figure}

In Table \ref{many_scores} we display the scores for our emulator variants. We achieve good performance with a $77.1\%$ $R^2$ score and an MSE of $0.16$. 
Using the correction method results per construction in no negative predictions, but increases the mass violation. The accuracy scores are not negatively affected by the correction operation. With the completion method, we achieve perfect mass conservation and a very slight worsening of the other metrics.
The mass regularization decreases the overall mass violation and decreases the mass biases for most cases. The positivity loss term decreases the negative fraction and negative mean by a large amount. A few examples are plotted in Figure \ref{four}. Overall a the architecture with an additional correction procedure could be a good choice to use in a GCM run, having the guarantee of no negative masses and good accuracy in the original units. The mass violation is relatively low for all cases.

Although the emulator performs well in an offline evaluation, it still remains to be shown how it performs when implemented in the GCM and run for multiple time steps. Good offline performance is not necessarily a  guarantee for good online performance \cite{gmd-13-2185-2020} in the case of convection parameterization, where a model crash could occur. In our case, it is likely that offline performance is a good indicator of online performance, as a global climate model crash is not expected because aerosols do not strongly affect large-scale dynamics.

\subsection{Runtime}

We conduct a preliminary runtime analysis by comparing the Python runtime for the emulator with the Fortran runtime of the original aerosol model. We take the time for one global time step, which is about 570,392 data points to predict. For the M7 model, the 31 vertical levels are calculated simultaneously and for the emulator, we predict the one time step at once, using a batch size of 571,392 and taking into account the time for transforming variables and moving them on and off the GPU. We use a single NVIDIA Titan V GPU and a single Intel Xeon 4108 CPU.  As shown in Table \ref{runtime} we can achieve a large speed-up of over 11,000x in a pure GPU setting. Including the time it takes to move the data from the CPU onto the GPU and back the acceleration is 64x compared to the original model. In case of no available GPU, the NN emulator is still 2.8 times faster. Here, further speed-ups will be achieved by using multiple CPUs, a smaller network architecture, and efficient implementation in Fortran.

\begin{table}[t]
\caption{Runtime comparison for the original M7 model and the NN emulator. NN pure GPU includes only the transformation of the variables and the prediction, whereas NN CPU-GPU-CPU also includes the time to transfer the data from the CPU to the GPU  and back to the CPU.}
\label{runtime}
\vskip 0.15in
\begin{center}
\begin{small}
\begin{sc}
\begin{tabular}{lcccr}
\toprule
Model & M7 & NN pure GPU & NN CPU-GPU-CPU & NN CPU  \\
\midrule
time (s)    & 5.781& 0.000517 &0.0897 & 2.042\\
speed-up & - & 11181.8 & 64.4 & 2.80\\
\bottomrule
\end{tabular}
\end{sc}
\end{small}
\end{center}
\vskip -0.1in
\label{overal_performance}
\end{table}

\section{Conclusion and Future Work}

To enable accurate climate forecasts aerosols need to be modeled with lower computational costs. This work shows how neural networks can be used to learn the mapping of an aerosol microphysics model and how simple physical constraints can be included in the model. Our neural models approximate the log-transformed tendencies excellently and the original units' tendencies well. On the test set, an overall regression coefficient of $77.1\%$ as well as an MSE of $16.1\%$ are achieved. Using a GPU we accomplish a large speed-up of 11,1181x compared to the original M7 model in a pure GPU setting, with the time to move from and to the CPU we are still significantly faster having a speed-up factor of 64. On a single CPUm the speed-up is 2.8x. By adding completion and correction mechanisms we can remove mass violation or negative predictions completely and make our emulator feasible for stable a GCM run. 

How much of a speed-up can be achieved in the end remains to be shown, when the machine learning model is used in a GCM run. Different versions of our emulator need to be run within the global climate model to show which setup performs the best online. A further step would be the combination of the different methods for physics constraining, to achieve both mass conservation and mass positivity at the same time.

%\paragraph{Acknowledgments}

%Bibliography
\bibliographystyle{unsrt}  
\bibliography{templateArxiv}  

\section{Supplementary Material}
\subsection{Modelled Variables}

Table \ref{variables} shows all the variables used for our emulator. Overall we consider 36 quantities, the first eight shown in the Table are only input variables and are not changed by M7/our emulator. Masses and concentrations of different aerosol species are both inputs and outputs for the model. Additionally, we output the water content of different aerosol size bins.

\begin{table}[b]
\caption{Modelled variables.}
%\vspace{-0.7in}
%\vskip -0.65in
\begin{center}
\begin{small}
\begin{sc}
\begin{tabular}{lccr}
\toprule
Variable & Unit & Input & Output  \\
\midrule
Pressure & $\mbox{Pa}$ &$\surd$ & \\
Temperature & K&$\surd$& \\
Rel. Humidity & - &$\surd$ & \\
Ionization Rate & - &$\surd$  & \\
cloud cover & - &$\surd$ & \\
Boundary layer & - &$\surd$ & \\
Forest fraction & - &$\surd$ & \\
H2SO4 prod. rate  & $cm^{-3}s^{-1}$ & $\surd$&  \\
H2SO4 mass    & $\mu g\  m^{-3}$ &$\surd$& $\surd$ \\
SO4 ns mass & $molec.\  m^{-3}$&$\surd$& $\surd$ \\
SO4 ks mass & $molec.\  m^{-3}$&$\surd$& $\surd$ \\
SO4 as mass & $molec.\  m^{-3}$&$\surd$& $\surd$ \\
SO4 cs mass & $molec.\  m^{-3}$&$\surd$& $\surd$ \\
bc ks mass & $\mu g\  m^{-3}$&$\surd$& $\surd$ \\
bc as mass & $\mu g\  m^{-3}$&$\surd$& $\surd$ \\
bc cs mass & $\mu g\  m^{-3}$&$\surd$& $\surd$ \\
bc ki mass & $\mu g\  m^{-3}$&$\surd$&$\surd$ \\
oc ks mass & $\mu g\  m^{-3}$&$\surd$& $\surd$ \\
oc as mass & $\mu g\  m^{-3}$&$\surd$& $\surd$ \\
oc cs mass & $\mu g\  m^{-3}$&$\surd$& $\surd$ \\
oc ki mass & $\mu g\  m^{-3}$&$\surd$& $\surd$ \\
du as mass & $\mu g\  m^{-3}$&$\surd$& $\surd$ \\
du cs mass &$\mu g\  m^{-3}$&$\surd$& $\surd$ \\
du ai mass & $\mu g\  m^{-3}$&$\surd$& $\surd$ \\
du ci mass & $\mu g\  m^{-3}$& $\surd$& $\surd$ \\
ns concentration&$ cm^{-3}$ &$\surd$& $\surd$ \\
ks concentration & $ cm^{-3}$ &$\surd$& $\surd$ \\
as concentration & $ cm^{-3}$ &$\surd$& $\surd$ \\
cs concentration & $ cm^{-3}$ &$\surd$& $\surd$ \\
ki concentration & $ cm^{-3}$ &$\surd$& $\surd$ \\
ai concentration & $ cm^{-3}$ &$\surd$& $\surd$ \\
ci concentration & $ cm^{-3}$ &$\surd$& $\surd$\\
ns water& $kg\ m^{-3} $& & $\surd$ \\
ks water & $kg\ m^{-3} $& & $\surd$ \\
as water& $kg\ m^{-3} $ & &$\surd$ \\
cs water& $kg\ m^{-3} $ & & $\surd$ \\
\bottomrule
\end{tabular}
\end{sc}
\end{small}
\end{center}
\vskip -0.1in
\label{variables}
\end{table}

\subsection{Logarithmic Transformation}

Training and evaluating on a logarithmic scale we can achieve much better scores than in the original units. On the test set, we achieve an $R^2$-score of $97.4\%$. The predictive quality on the logarithmic scale though does not transfer to the original scale, as we can see in the scores in Table \ref{many_scores2}. 

\begin{table}[htb]
\tabcolsep=0pt%
{\caption{Test metrics for different architectures and transformations. Standard refers to learning with the standard transformation, log-scaled to the learning on logarithmitically transformed values. Base means the usage of only the base NN, correct adds the correction procedure and complete the completion procedure. Mass reg. and positivity reg. include the regularization terms. Best scores are bold and second best scores are highlighted in blue .\label{tab222}}}
{
\begin{tabular*}{\textwidth}{@{\extracolsep{\fill}}lccccccccc@{}}\toprule%
 & \multicolumn{5}{@{}c@{}}{{Standard}}& \multicolumn{4}{@{}c@{}}{{Log-Scaled}}
 \\\cmidrule{2-6}\cmidrule{7-10}%
{Architecture} & {Base} & {+Correct} & {+Complete} & {+Mass Loss} & {+Positivity Loss} &
{Base} & {+Correct} & {+Complete} & {+Mass Loss}\\\midrule
{R2}&\textcolor{blue}{$0.763$}&\textbf{0.771} &$0.738$&0.730 &0.709 &$-$& $-$& $-$&$-$\\
{R2 log}& $-$ &$-$&$-$ & $-$ & $-$ &\textcolor{blue}{$0.974$}& {$0.974$}& {0.959}&\textbf{0.984}\\

{RMSE}&\textcolor{blue}{0.402 }&\textbf{0.401} &0.403&0.433 & 0.459&115431& 115431&260706&21057757\\

{Mass Bias SO4}&$1.1\cdot 10^{-5}$ &$8.5\cdot 10^{-5}$ &\textbf{0.00} & \textcolor{blue}{$8.6\cdot 10^{-6}$}& $1.0\cdot 10^{-3}$&$1.2\cdot 10^{0}$ & $1.2\cdot 10^{0}$& \textbf{0.00} &$1.1\cdot 10^{-3}$\\
{Mass Bias BC}&{$3.8\cdot 10^{-5}$} &$1.4\cdot 10^{-4}$ &\textbf{0.00}& \textcolor{blue}{$3.4\cdot 10^{-5}$}&$3.6\cdot 10^{-4}$ &$2.6\cdot 10^{-4}$& $2.6\cdot 10^{-4}$& \textbf{0.00} &$3.2\cdot 10^{-4}$\\
{Mass Bias OC}&{$3.3\cdot 10^{-5}$} &$6.0\cdot 10^{-5}$&\textbf{0.00}& \textcolor{blue}{$1.1\cdot 10^{-5}$}& $6.4\cdot 10^{-4}$&$3.8\cdot 10^{-4}$& $1.4\cdot 10^{-4}$& \textbf{0.00} &$8.0\cdot 10^{-5}$\\
{Mass Bias DU}&{$1.0\cdot 10^{-6}$}&$3.9\cdot 10^{-5}$ &\textbf{0.00}& \textcolor{blue}{$2.8\cdot 10^{-7}$}& $1.5\cdot 10^{-5}$&$1.0\cdot 10^{-3}$& $2.2\cdot 10^{-4}$& \textbf{0.00} &$8.5\cdot 10^{-5}$\\
{Mass Violation}&$3.7\cdot 10^{-4}$ &$1.1\cdot 10^{-3}$ &\textbf{0.00}& \textcolor{blue}{$1.4\cdot 10^{-4}$} & $2.4\cdot 10^{-3}$&$4.0\cdot 10^{2}$ & $4.0\cdot 10^{2}$& \textbf{0.00} &$3.2\cdot 10^{-1}$\\
{Neg. Fraction}&0.134 &\textbf{0.00} &0.146 & 0.144& 0.0894 &{0.0807}&\textbf{0.00} & 0.0812&\textcolor{blue}{0.061}\\
{{Neg. Mean}}&0.00151&\textbf{0.00} &0.00170 & {0.00169}& \textcolor{blue}{0.000081} &0.00921&\textbf{0.00} &0.108&0.00115\\

\label{many_scores2}
\end{tabular*}%
}
\end{table}

\subsection{Other ML Approaches}

Apart from a neural network approach, we investigated other ML models to emulate the M7 module. We looked at linear regression (LR) and different ensemble methods such as a random forest regressor t (RF) and a gradient boosting model (GB). In Table \ref{other ML approaches} we report the test scores for linear regression, a random forest, and a gradient boosting approach. Being linear combinations of training points, linear regression and random forest both automatically conserve mass. Overall the random forest shows the best accuracy among these models. We note that the random forest performs worse than the neural network in terms of predictive accuracy, but could still be considered for future work because of the mass conserving property.

\begin{table}[t]
\tabcolsep=0pt%
{\caption{Test metrics for different other ML models.}}
{
\begin{tabular*}{\textwidth}{@{\extracolsep{\fill}}lcccccccccc@{}}\toprule%
 & \multicolumn{3}{@{}c@{}}{{Accuracy}}& \multicolumn{5}{@{}c@{}}{{Mass Conservation}}& \multicolumn{2}{@{}c@{}}{{Positivity}}
 \\\cmidrule{2-4}\cmidrule{5-9}\cmidrule{10-11}%
{Architecture} & {R2} & {R2 log} & {RMSE} & {Bias SO4}&{Bias BC} & {Bias OC} &
{Bias DU} & {RMSE} & {Neg. Fraction} & {Neg. Mean}\\\midrule
{LR}&+2.10e-1& -4.19e+0 & +6.95e-1& -4.21e-18&-1.03e-16 & -1.08e-16& +1.03e-17& +4.16e-16&+1.92e-1&-1.32e-2\\
{RF}&+7.07e-1& -7.30e-1 & +4.66e-1& +9.90e-17&-4.78e-16 & -6.19e-17& -9.27e-19&+3.85e-16&+1.31e-1&-7.28e-4\\
{GB}&+6.66e-1& -2.95e-1 & +4.76e-1& +1.90e-5&+1.75e-5 & +1.45e-5& -8.56e-5& +1.14e-2&+1.27e-1&-1.25e-3\\

\label{other ML approaches}
\end{tabular*}%
}
\end{table}

\subsection{Network Architectures and Losses}

\begin{figure}[ht]
\vskip 0.2in
\begin{center}
\centerline{\includegraphics[width=14cm]{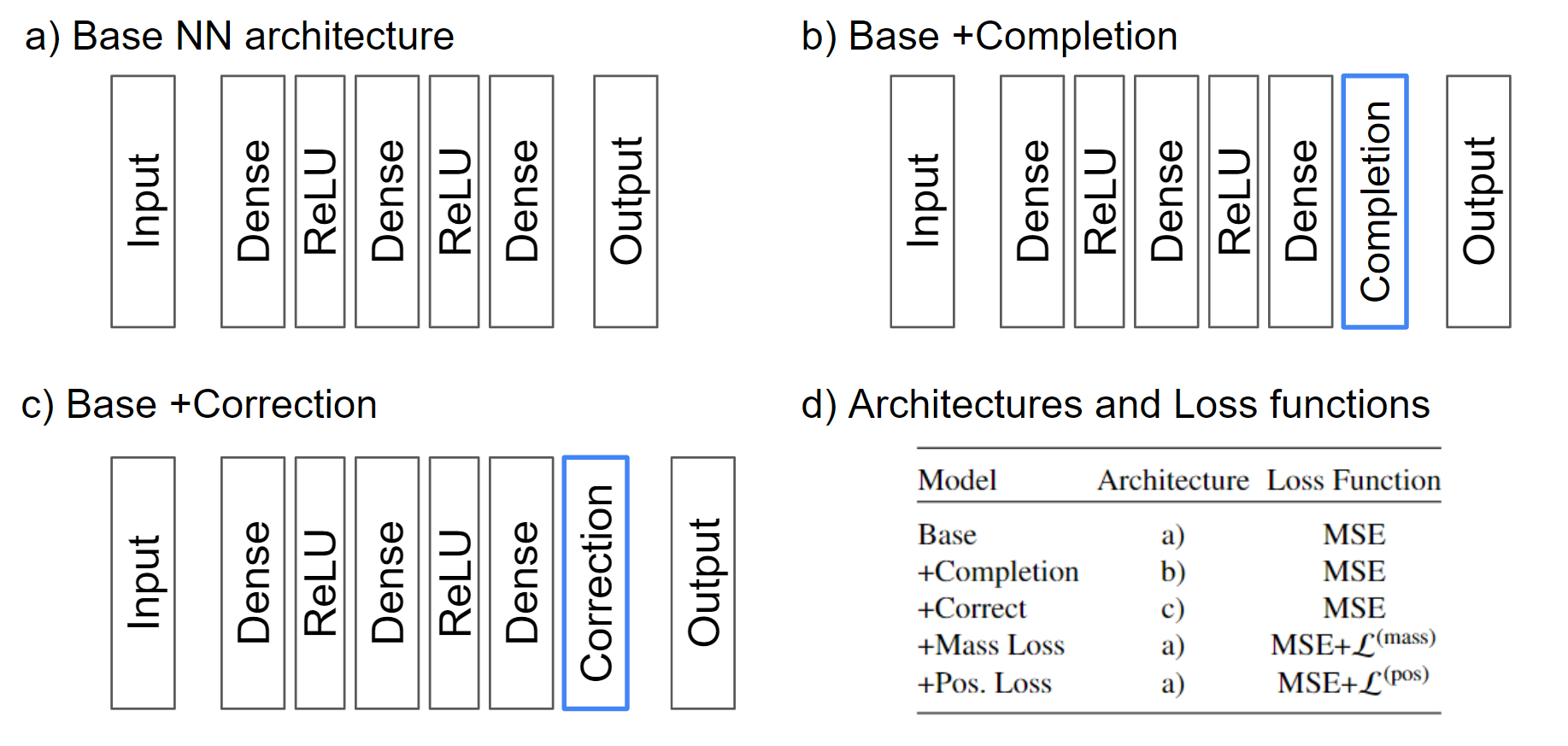}}
\caption{Different architectures and loss functions used for comparison.}
\label{archi}
\end{center}
\end{figure}

The network architectures used all build on the same base neural network containing two hidden layers and ReLU activations. For enforcing hard constraints we add an additional layer at the end, either the correction layer for the inequality constraint of positive masses or the completion layer of the equality constraint for mass conservation. Schematics and the combination of architectures and loss functions are shown in Figure \ref{archi}.

\subsection{Activation Functions}

We explored different activation functions for our neural architecture, Sigmoid, Tanh, ReLU, and Leaky ReLU. We evaluated their performance (see Table \ref{activation}) on the validation set and find that the ReLU-based activations result in better MSE and $R^2$-scores.

\begin{table}[t]
\caption{Validation scores for different activation functions.}
\label{sample-table}
\vskip 0.15in
\begin{center}
\begin{small}
\begin{sc}
\begin{tabular}{lcr}
\toprule
Activation Function & MSE & $R^2$  \\
\midrule
sigmoid     &0.442 & 74.2\\
tanh & 0.446&73.4 \\
relu & 0.395&78.9 \\
leaky relu &0.395&78.7 \\
\bottomrule
\end{tabular}
\end{sc}
\end{small}
\end{center}
\vskip -0.1in
\label{activation}
\end{table}

\subsection{Parameters for Regularization}

The neural network's success is very sensitive to changes in the parameters of both mass and positivity regularizers. The parameters we found to work the best in this scenario are $\alpha = [10^{-7}, 2\cdot 10^{4} ,2\cdot 10^{3} ,10^{-1} ]$ and $\beta = [10^{-11}, 10^{7},  10^{7},10^{3},10^{-8},10^{1}]$ for the training with linear transformed values and $\alpha = [10^{-8}, 10^3, 10^{4}, 10^5]$ for log-transformation. $\beta$ was tuned for the different variable groups (SO4, BC, OC, DU, NUM, WAT) and not each variable independently, which could lead to further improvement.

\subsection{Transformation for Constraining}

For clarity and simplicity, we omit the fact that we need to back-transform the variables before calculating masses and negative fraction in the main paper. The more technically precise formulation is introduced here. Let $\mu_x,\sigma_x\in \mathbb{R}^{32}$ be the mean and standard deviation vectors over all training input data and $\mu_y, \sigma_y \in \mathbb{R}^{28}$ the mean over all training output data. Then we set the function $g: \mathbb{R}^{28}\mapsto\mathbb{R}^{28}$ as the back transformation for the output data, $g(y) = y*\sigma_y+\mu_y$ and $h: \mathbb{R}^{32}\mapsto\mathbb{R}^{32}$ as the back transformation for the input data, $h(x) = x*\sigma_x+\mu_x$. With that, the loss terms are defined as:

\begin{equation}\label{loss2}
    \mathcal{L}^{\text{(mass)}} (y_\theta) := \sum_{s \in S} \alpha_s|\sum_{i \in I_s} g(y_\theta^{(i)}) |,
\end{equation}

for the mass conservation and

\begin{equation}
    \mathcal{L}^{\text{(pos)}} (y_\theta) := \sum_{i=0}^{n}\beta_i ReLU(- (g(y_\theta^{(i)})+h(x_i)))^2
\end{equation}
 for the positivity term.
 
 The same applies for hard-constraining. The equations for that are:
 
 \begin{equation}
     y_\theta^{(i)} = \text{ReLU}((g(\tilde{y}_\theta^{(i)})+h(x_i))) - h(x_i)
\end{equation}
 and 
\begin{equation}
     y_\theta^{(j)} = (-\sum_{i \in I_s\setminus \{ j \}} g(y_\theta^{(i)}) - \mu_{y_j})/\sigma_{y_j}.
\end{equation}
 
\subsection{Metrics Development}

\begin{figure}[ht]
\vskip 0.2in
\begin{center}
\centerline{\includegraphics[width=16cm]{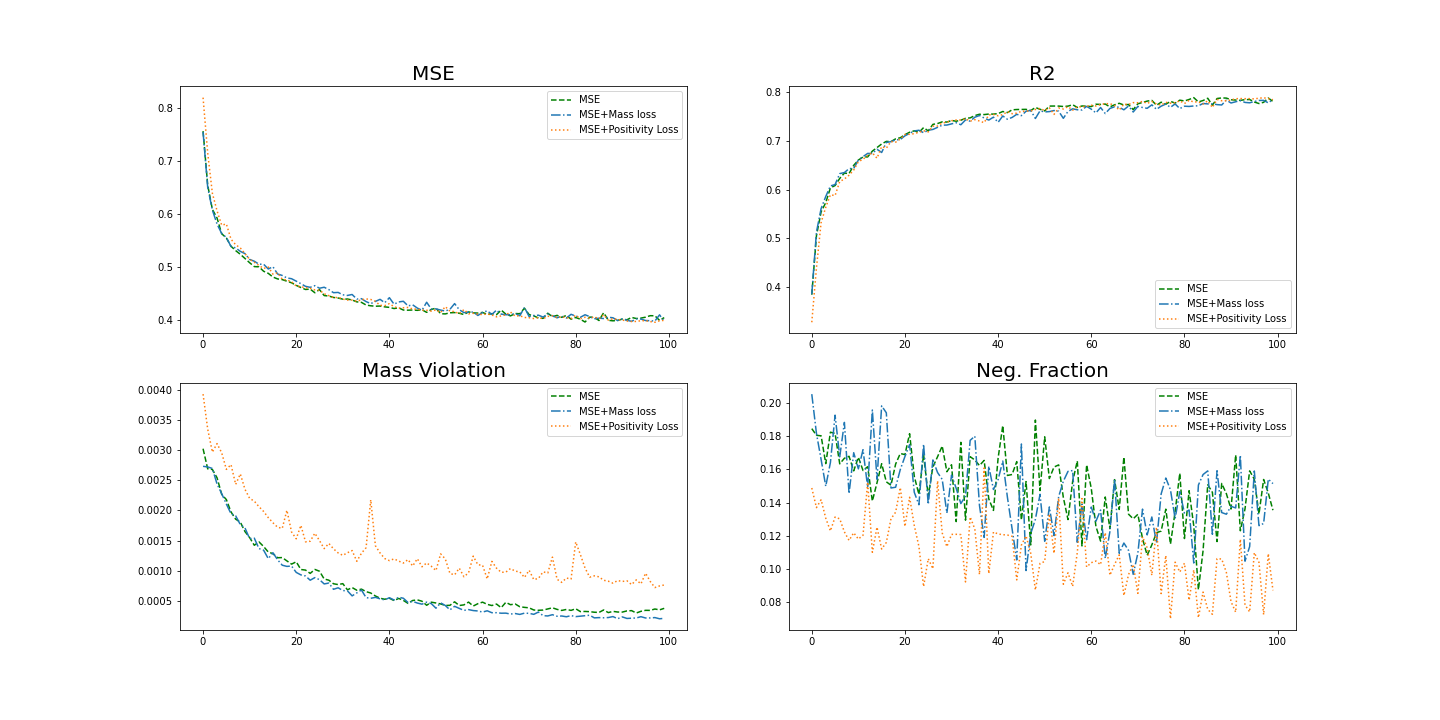}}
\caption{Development of MSE, $R^2$-score, mass conservation violation and negative fraction over the course of training epochs, measured on the validation set.}
\label{epochs}
\end{center}
\end{figure}

In Figure \ref{epochs} we show the development of MSE, $R^2$-score, average mass conservation violation, and overall negative fraction for different losses throughout training. We can observe that the accuracies expressed in MSE and $R^2$-score develop very similar for all loss types. Adding the mass violation term we can see an improvement in mass conservation, especially for later layers, adding the positivity term we can see a decrease in negative fraction for all the epochs. It is also noticeable, that the negative fraction shows high variability between different epochs.

\subsection{Results for Classification}

To enable log-transformed learning we need to combine our regression neural network with a classification network. The classification network predicts whether a tendency is positive or negative. The scores for this binary classification task can be found in Table \ref{class_scores}. Note that this classification problem is highly imbalanced and could be improved by using techniques for this case.

\begin{table}[t]
\tabcolsep=0pt%
{\caption{Scores for classification neural network, calculated on test set.}}
{
\begin{tabular*}{\textwidth}{@{\extracolsep{\fill}}lccccccccccc@{}}\toprule
{Variable} & {SO4 NS} & {SO4 KS} & {SO4 AS} & {BC KS}&{BC AS} & {OC KS}&{OC AS}& {DU AS} &
{NUM NS} & {NUM KS} & {NUM AS}\\\midrule
{Accuracy}&0.986& 0.962& 0.999&0.960&0.999 & 0.962& 0.999& 0.999&0.989&0.984&0.959\\
{Precision}&0.954& 0.931 & 0.999& 0.914&0.999 & 0.908& 0.999&1.000&0.972&0.755&0.919\\
{Recall}&0.990& 0.988 & 1.000&0.972& 1.000&0.968 &1.000& 1.000& 0.989& 0.844&0.768\\

\label{class_scores}
\end{tabular*}%
}
\end{table}

\begin{table}[t]
\tabcolsep=0pt%
{\caption{$R^2$ scores on the test set for each variable. For our neural network that learns the log-transformed tendencies we consider the $R^2$ score on the log-transformed values, for the neural network trained on standardized values we consider the $R^2$ in original units.\label{tab22}}}
{
\begin{tabular*}{\textwidth}{@{\extracolsep{\fill}}lccccccccccccc@{}}\toprule%
 & \multicolumn{5}{@{}c@{}}{{SO4}}& \multicolumn{4}{@{}c@{}}{{Black Carbon}} & \multicolumn{4}{@{}c@{}}{{Organic Carbon}}
 \\\cmidrule{2-6}\cmidrule{7-10}\cmidrule{11-14}%
{Variable} & {H2SO4} & {NS} & {KS} & {AS} & {CS} &
{KS} & {AS} & {CS} & {CI}&{KS} & {AS} & {CS} & {CI}\\\midrule
{Log. ($R^2$) }& 0.999 & 0.993 & 0.981&0.993&0.992&0.979&0.995&0.990&0.997&0.984&0.997&0.994&0.998\\
{Stand. ($R^2$)}& 0.983& 0.877&0.652&0.875&0.292&0.837&0.844&0.000&0.867&0.616&0.669&0.200&0.710 \\

\label{scores_var1}
\end{tabular*}%
}
\vskip -0.2in
\end{table}

\begin{table}[t]
\tabcolsep=0pt%
{\caption{Same as table above for additional variables.}}
{
\begin{tabular*}{\textwidth}{@{\extracolsep{\fill}}lcccccccccccccccc@{}}\toprule%
 & \multicolumn{4}{@{}c@{}}{{Dust}}& \multicolumn{7}{@{}c@{}}{{Number particles}}&\multicolumn{4}{@{}c@{}}{{Water content}}
 \\\cmidrule{2-5}\cmidrule{6-12}\cmidrule{13-16}%
{Variable} & {AS} & {CS} & {AI} & {CI} & {NS} &
{KS} & {AS} & {CS} & {KI}&{AI} & {CI} & {NS} & {KS} & {AS} & {CS}\\\midrule
{Log. ($R^2$)}&0.995&0.996 &0.995  & 0.996&0.993&0.997&0.998&0.997&0.997&0.977&0.983&0.979&0.982 &0.979&0.982\\
{Stand. ($R^2$)}& 0.872& 0.966&0.945&0.966&0.900&0.663&0.874&0.932&0.949&0.936&0.940&0.979&0.978&0.928&0.954 \\

\label{scores_var2}
\end{tabular*}%
}
\end{table}

\subsection{Individual Scores and Plots}

\begin{figure*}[ht]
\vskip 0.2in
\begin{center}
\centerline{\includegraphics[width=14cm]{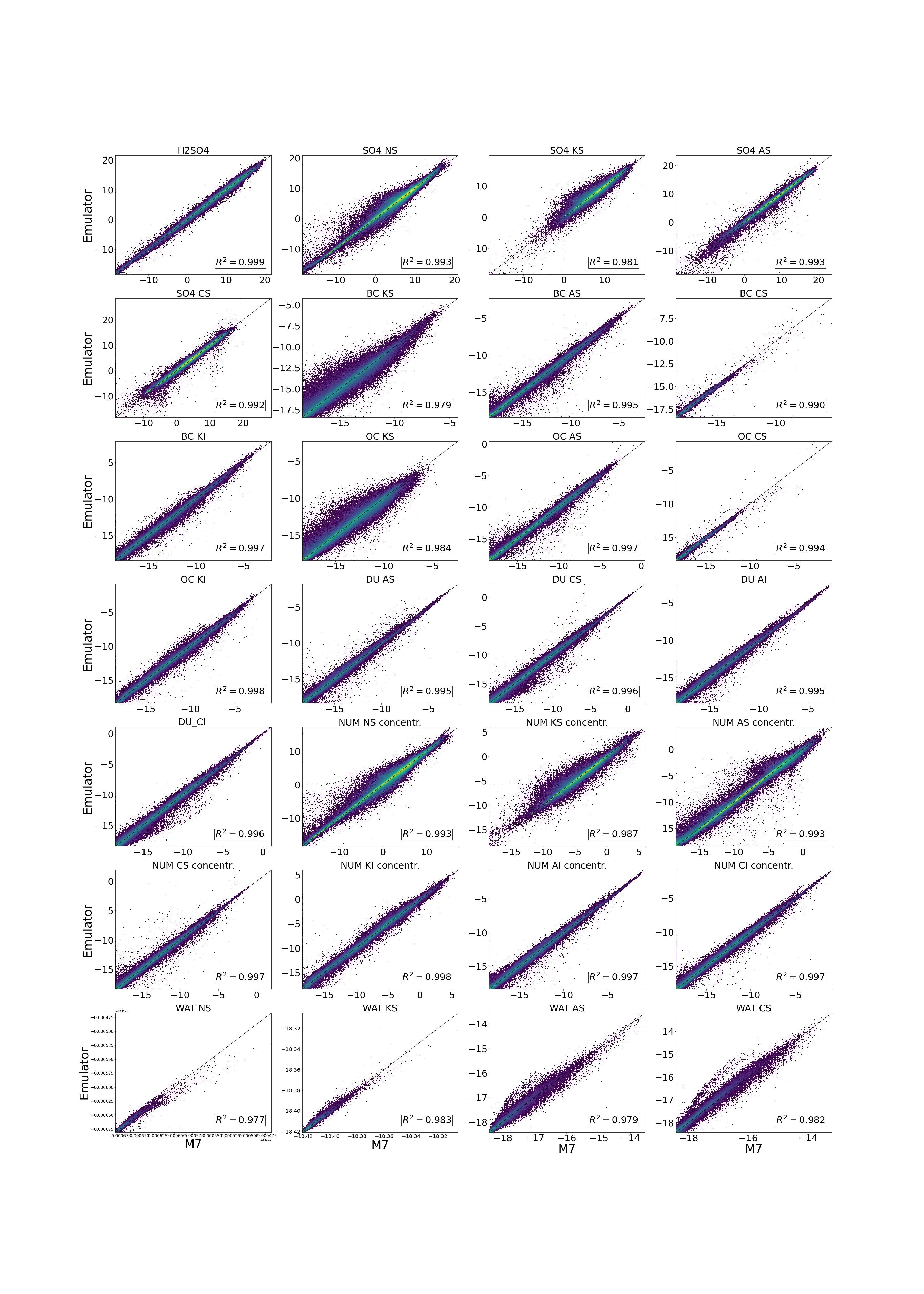}}
\caption{Predicted test values by emulator versus true values from M7 model. Trained on logarithmically transformed quantities and plotted in the logarithmic scale}
\label{all_log}
\end{center}
\end{figure*}

\begin{figure*}[ht]
\vskip 0.2in
\begin{center}
\centerline{\includegraphics[width=14cm]{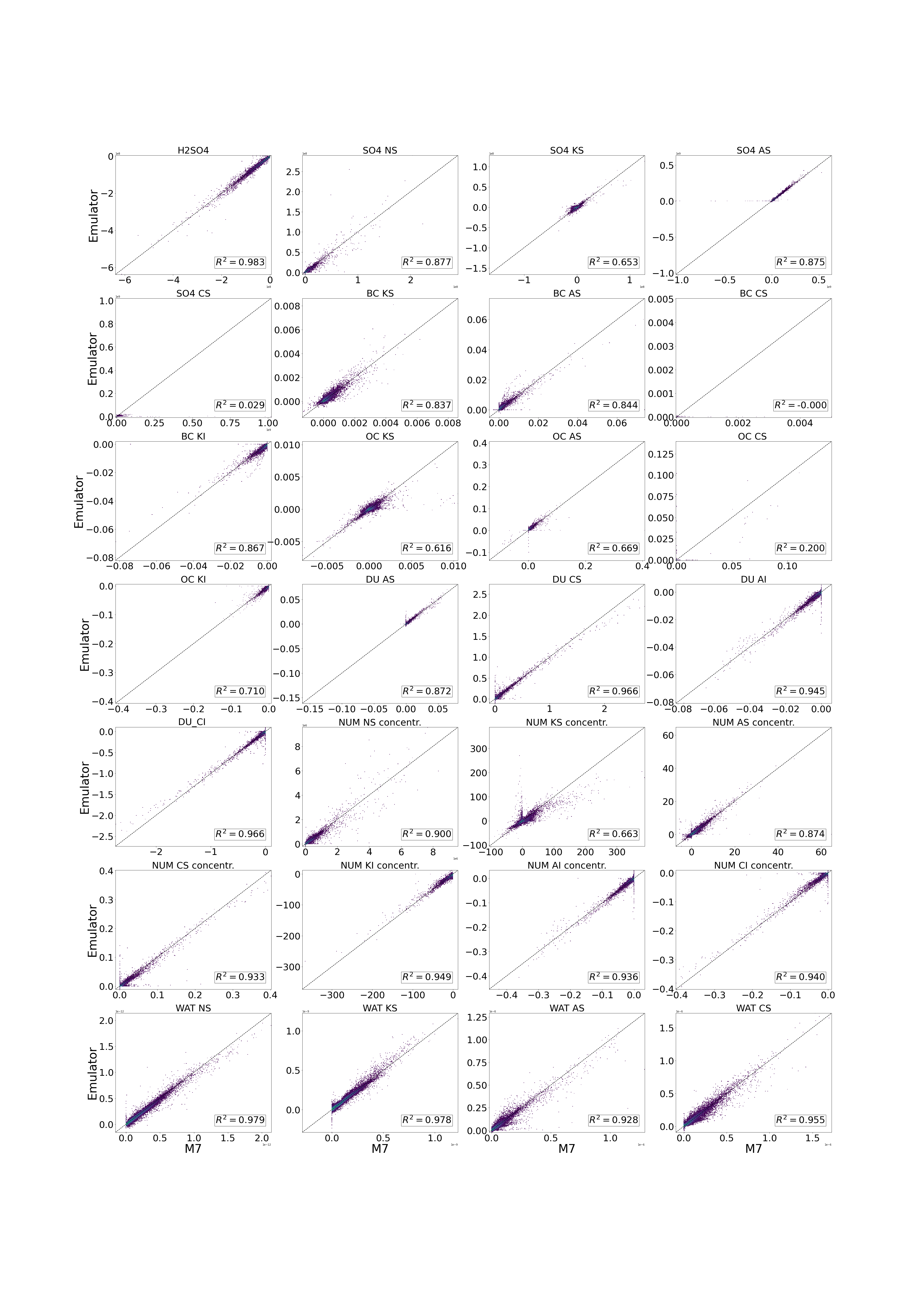}}
\caption{Predicted test values by emulator versus true values from M7 model. Trained on standardized quantities and plotted standardized}
\label{all}
\end{center}
\end{figure*}

In Table \ref{scores_var1} and \ref{scores_var2} we present the $R^2$ scores for all predicted variables individually, using the base neural network for logarithmic transformed or standardized values. Figure \ref{all_log} and \ref{all} show plots for all predicted variables. 

\end{document}